\documentclass[conference]{IEEEtran}
\IEEEoverridecommandlockouts
\usepackage{cite}
\usepackage{url}

\usepackage[ruled]{algorithm2e}
\usepackage{amsmath,amssymb,amsfonts}
\usepackage{algorithmic}
\usepackage{graphicx}
\usepackage{textcomp}
\usepackage{xcolor}
\usepackage{bm}
\usepackage{multirow}
\usepackage{booktabs}
\usepackage[figuresright]{rotating}
\UseRawInputEncoding
\def\BibTeX{{\rm B\kern-.05em{\sc i\kern-.025em b}\kern-.08em
    T\kern-.1667em\lower.7ex\hbox{E}\kern-.125emX}}
\begin{document}

\title{Accelerating differential evolution algorithm with Gaussian sampling based on estimating the convergence points\\}

\author{\IEEEauthorblockN{1\textsuperscript{st} Rui Zhong}
\IEEEauthorblockA{\textit{Graduate School of Information Science and Technology} \\
\textit{Hokkaido University}\\
Sapporo, Japan \\}
\and
\IEEEauthorblockN{2\textsuperscript{nd} Masaharu Munetomo}
\IEEEauthorblockA{\textit{Information Initiative Center} \\
\textit{Hokkaido University}\\
Sapporo, Japan \\}
}
\maketitle

\begin{abstract}
In this paper, we propose a simple strategy for estimating the convergence point approximately by averaging the elite sub-population. Based on this idea, we derive two methods, which are ordinary averaging strategy, and weighted averaging strategy. We also design a Gaussian sampling operator with the mean of the estimated convergence point with a certain standard deviation. This operator is combined with the traditional differential evolution algorithm (DE) to accelerate the convergence. Numerical experiments show that our proposal can accelerate the DE on most functions of 28 low-dimensional test functions on the CEC2013 Suite, and our proposal can easily be extended to combine with other population-based evolutionary algorithms with a simple modification. 
\end{abstract}


\section{Introduction} \label{sec:1}

As an important part of artificial intelligence, evolutionary computation has achieved great success in solving continuous\cite{Ahandani:10}, large-scale\cite{Yang:07}, constraint\cite{Wang:15}, and multi-objective optimization problems\cite{Deb:11} in the past few decades. However, according to the No Free Lunch Theorem (NFLT)\cite{Wolpert:97}, there is no optimization algorithm that can solve all optimization problems perfectly. NFLT proves that the average performance of any pair of algorithms $A$ and $B$ is identical on all possible problems. Therefore, if an algorithm performs well on a certain class of problems, it must pay for that with performance degradation on the remaining problems, since this is the only way for all algorithms to have the same performance on average across all functions. This hypothesis has a great impact on the scientific community for the notion that there is no universal optimizer. In addition, a single algorithm often cannot balance the exploration and the exploitation well. For several reasons, hybrid algorithms or memetic algorithms appears, the hybrid algorithm was first proposed to solve the Traveling Salesman Problem (TSP) by modifying the genetic algorithm with a local search operator\cite{Michael:91}. However, NFLT also limits the hybrid algorithms with identical average performance on all possible problems, thus the features of the problem become the starting point for designing a suitable optimization algorithm. 

Since hybrid algorithms were not proposed as specific optimization algorithms, but as a broad class of algorithms inspired by the diffusion of ideas and composed of multiple existing operators, the community started showing increasing attention toward these algorithmic structures as a general guideline for addressing specific problems.

In addition, many studies\cite{Abdolmaleki:17,Fides:22} show that it is a promising strategy to find trusting regions or elites in the fitness landscape, which can be fully applied to guide the direction of evolution well. Murata et al\cite{Noboru:15}. first proposed that a mathematical method could be used to calculate the global optimum using the information of two subsequent generations. Yu et al\cite{Yu:19}. proposed that in the differential evolution algorithm, the differential vector from the parent individual to its offspring (or the vector from an individual with poor fitness to an individual with higher fitness) is defined as the moving vector, and the convergence point is estimated according to the moving vector.

In this paper, we employ a simple strategy to approximately estimate the convergence point, which is roughly estimated by the averaging strategy and weighted averaging strategy of the elite sub-population. Besides, after each generation optimized by DE, we apply a Gaussian sampling operator with the mean of the estimated convergence point. In experiment design, we evaluate our proposal with 28 benchmark functions from the CEC 2013 Suite\cite{Liang:13}. Finally, we summarize our work and provide some open topics for future discussions.

The rest of the paper is organized as follows. Section \ref{sec:2} includes preliminaries and related works. Section \ref{sec:3} provides a detailed description of our proposal. Section \ref{sec:4} provides the numerical experiments and experiments analysis. Finally, Section \ref{sec:5} concludes the paper and shows the future directions.

\section{Preliminaries and Related works} \label{sec:2}
In this section, we first introduce the preliminaries of this work, including the Random Search (RS), Genetic Algorithm (GA), Differential Evolution Algorithm (DE), Evolution Strategy (ES), and Particle Swarm Optimization (PSO). Finally, we introduce the original estimation of a convergence point.

\subsection{Preliminaries}
\subsubsection{RS}
RS was proposed by Rastrigin in 1963, and an early introduction to RS with basic mathematical analysis was given in the paper \cite{Rastrigin:63}. The principle of RS is to iteratively move to better positions in the search space, these positions are randomly sampled around the current optimum. RS is an optimization method that does not require the gradients of the fitness landscape, so RS can be employed for discrete or differentiable problems. The Pseudocode of RS is shown in Algorithm \ref{alg:1}.
\begin{algorithm}
	\label{alg:1}
	\DontPrintSemicolon
	\SetAlgoLined
	\KwIn {${\rm Dimension}:D;{\rm Search \ space}:S;{\rm Generation}:T$}
	\KwOut {${\rm Best \ solution}: E$}
	\SetKwFunction{FRS}{\textbf{RS}}
	\SetKwProg{Fn}{Function}{:}{}
	\Fn{\FRS{$D, S, T$}}{
		$t \gets 0$\;
		$\blacktriangleright$ (Solution initialization) \;
		$P_t \gets \textbf{Initial}(D, S)$ \;
		$FP \gets \textbf{Evaluate}(P_t)$\;
		$E \gets P_t$\;
		\While{$t < T$ {\rm and not convergence}}{
			$O_t \gets \textbf{Sampling}(P_t)$ \;
			$FO \gets \textbf{Evaluate}(O_t)$\;
			\If{$FO > FP$}{
				$P_t \gets O_t$ \;
				$E \gets O_t$\;
			}
			$t \gets t+1 $\;
		}
		$\textbf{return} \ E$
	}
	\caption{RS}
\end{algorithm}

\subsubsection{GA}
GA\cite{Holland:92} simulates the biological evolution process of chromosomes with selection, crossover, and mutation. Chromosomes present problem solutions and are evaluated based on the fitness function to select parents. Selection is an important process for choosing parents to produce the offspring and can affect the convergence of GA, Mutation increases the diversity of the population by randomly modifying the genes in the chromosome with a certain probability. The Pseudocode of GA is shown in Algorithm \ref{alg:2}
\begin{algorithm}
	\label{alg:2}
	\DontPrintSemicolon
	\SetAlgoLined
	\KwIn {${\rm Dimension}:D;{\rm Search \ space}:S;{\rm Population \ size}:PS;{\rm Generation}:T$}
	\KwOut {${\rm Best \ solution}: E$}
	\SetKwFunction{FGA}{\textbf{GA}}
	\SetKwProg{Fn}{Function}{:}{}
	\Fn{\FGA{$D, S, PS, T$}}{
		$t \gets 0$\;
		$\blacktriangleright$ (Population initialization) \;
		$P_t \gets \textbf{Initial}(D, S, PS)$ \;
		$F_t \gets \textbf{Evaluate}(P_t)$\;
		$E \gets \textbf{bestIndividual}(P_t, F_t)$\;
		\While{$t < T$ {\rm and not convergence}}{
			$\blacktriangleright$ (Selection, Crossover and Mutation) \;
			$O_t \gets \textbf{Selection}(P_t, F_t)$\;
			$O_t \gets \textbf{Crossover}(O_t)$\;
			$P_{t+1} \gets \textbf{Mutation}(O_t)$\;
			$F_{t+1} \gets \textbf{Evaluate}(P_{t+1})$\;
			$E \gets \textbf{bestIndividual}(P_{t+1}, F_{t+1})$\;
			$t \gets t+1 $\;
		}
		$\textbf{return} \ E$
	}
	\caption{GA}
\end{algorithm}

\subsubsection{DE}
DE was first proposed by Storn and Price in 1995\cite{Storn:96}. DE has been applied to solve many complex optimization problems, and the procedure is similar to the GA, DE mainly includes mutation, crossover, and selection. The Pseudocode of DE is shown in Algorithm \ref{alg:3}.
\begin{algorithm}
	\label{alg:3}
	\DontPrintSemicolon
	\SetAlgoLined
	\KwIn {${\rm Dimension}:D;{\rm Search \ space}:S;{\rm Population \ size}:PS;{\rm Generation}:T$}
	\KwOut {${\rm Best \ solution}: E$}
	\SetKwFunction{FDE}{\textbf{DE}}
	\SetKwProg{Fn}{Function}{:}{}
	\Fn{\FDE{$D, S, PS, T$}}{
		$t \gets 0$\;
		$\blacktriangleright$ (Population initialization) \;
		$P_t \gets \textbf{Initial}(D, S, PS)$ \;
		$F_t \gets \textbf{Evaluate}(P_t)$\;
		$E \gets \textbf{bestIndividual}(P_t, F_t)$\;
		\While{$t < T$ {\rm and not convergence}}{
			$\blacktriangleright$ (Selection, Crossover and Mutation) \;
			$O_t \gets \textbf{Mutation}(P_t, F_t)$\;
			$O_t \gets \textbf{Crossover}(O_t)$\;
			$F_{t+1} \gets \textbf{Evaluate}(O_t)$\;
			$P_{t+1} \gets \textbf{Selection}(O_t, F_{t+1})$\;
			$E \gets \textbf{bestIndividual}(P_{t+1}, F_{t+1})$\;
			$t \gets t+1 $\;
		}
		$\textbf{return} \ E$
	}
	\caption{DE}
\end{algorithm}

Mutation is an essential process in DE. We briefly introduce 2 common mutation strategies:
\begin{equation}
	\begin{aligned}
		\textbf{DE/rand/1}:V_i(g)=X_{p1}(g)+F \cdot (X_{p2}(g)-X_{p3}(g)) \\
		\textbf{DE/best/1}:V_i(g)=X_{best}(g)+F \cdot (X_{p1}(g)-X_{p2}(g)) 
		 \nonumber
	\end{aligned}
\end{equation}
$X_{pi}(g)$ is the different individuals randomly selected from the current population, $X_{best}(g)$ is the best individual in the current population. $F$ is the scaling factor. The differential vectors between individuals are calculated in the mutation is the origin of the name of DE.

\subsubsection{ES}
ES was proposed in 1963\cite{Slobodkin:64}, as an optimization algorithm, ES imitates the mechanism of biological evolution. Under the hypothesis that no matter what changes occur in genes, the results or traits always follow a Gaussian distribution with zero mean and a certain variance. Although ES employs mutation and crossover to generate offspring which is similar to GA, ES emphasizes the phenotype rather than the genotype, and ES also applies the real coding instead of binary coding in GA. The Pseudocode of ES is shown in Algorithm \ref{alg:4}.
\begin{algorithm}
	\label{alg:4}
	\DontPrintSemicolon
	\SetAlgoLined
	\KwIn {${\rm Dimension}:D;{\rm Search \ space}:S;{\rm Population \ size}:PS;{\rm Generation}:T$}
	\KwOut {${\rm Best \ solution}: E$}
	\SetKwFunction{FES}{\textbf{ES}}
	\SetKwProg{Fn}{Function}{:}{}
	\Fn{\FES{$D, S, PS, T$}}{
		$t \gets 0$\;
		$\blacktriangleright$ (Population initialization) \;
		$P_t \gets \textbf{Initial}(D, S, PS)$ \;
		$FP_t \gets \textbf{Evaluate}(P_t)$\;
		$E \gets \textbf{bestIndividual}(P_t, F_t)$\;
		\While{$t < T$ {\rm and not convergence}}{
			$\blacktriangleright$ (Selection, Crossover and Mutation) \;
			$O_t \gets \textbf{Selection}(P_t)$\;
			$O_t \gets \textbf{Crossover}(O_t)$\;
			$O_t \gets \textbf{Mutation}(O_t)$\;
			$FO_t \gets \textbf{Evaluate}(O_t)$\;
			$P_{t+1}, FP_{t+1} \gets \textbf{Offspring}(O_t, FO_t, P_t, FP_t)$\;
			$E \gets \textbf{bestIndividual}(P_{t+1}, FP_{t+1})$\;
			$t \gets t+1 $\;
		}
		$\textbf{return} \ E$
	}
	\caption{ES}
\end{algorithm}

\subsubsection{PSO}
PSO simulates the behavior of bird flocking, fish schooling, and swarming theory to realize the optimization process\cite{Kennedy:95}. The original version of PSO as developed by the authors comprises a very simple concept, and paradigms can be implemented in a few lines of computer code. It requires only primitive mathematical operators and is computationally inexpensive in terms of both memory requirements and speed. The Pseudocode of PSO is shown in Algorithm \ref{alg:5}.
\begin{algorithm}
	\label{alg:5}
	\DontPrintSemicolon
	\SetAlgoLined
	\KwIn {${\rm Dimension}:D;{\rm Search \ space}:S;{\rm Population \ size}:PS;{\rm Generation}:T$}
	\KwOut {${\rm Best \ solution}: gbest$}
	\SetKwFunction{FPSO}{\textbf{PSO}}
	\SetKwProg{Fn}{Function}{:}{}
	\Fn{\FPSO{$D, S, PS, T$}}{
		$t \gets 0$\;
		$\blacktriangleright$ (Initialize the position and velocity) \;
		$X_t,V_t \gets \textbf{Initial}(D, S, PS)$ \;
		\For{$i=0 \ to \ PS$}{
			$pbest_i \gets X_{t,i}$  \;
		}
		$F_t \gets \textbf{Evaluate}(X_t)$\;
		$gbest \gets \textbf{bestIndividual}(X_t, F_t)$\;
		\While{$t < T$ {\rm and not convergence}}{
			$\blacktriangleright$ (Optimization) \;
			\For{$i=0 \ to \ PS$}{
				$X_{t+1,i} \gets \textbf{Update}(X_{t,i}, V_{t,i})$ \;
				$V_{t+1,i} \gets \textbf{Update}(V_{t,i}, pbest_i, gbest)$ \;
			}
			$F_{t+1} \gets \textbf{Evaluate}(X_{t+1})$\;
			\For{$i=0 \ to \ PS$}{
				\If{$F_{t+1, i} > F_{pbest, i}$}{
					$ pbest_i\gets X_{t+1, i}$\;
				}
				\If{$F_{pbest, i} > F_{gbest, i}$}{
					$ gbest_i\gets pbest_i$\;
				}
			}
			$t \gets t+1 $\;
		}
		$\textbf{return} \ gbest$
	}
	\caption{PSO}
\end{algorithm}
We update the position and velocity by Eq (\ref{eq:4}) in lines 13 and 14 of the Algorithm \ref{alg:5}
\begin{equation}
	\begin{aligned}
		\label{eq:4}
		x_{id}^{k+1}=x_{id}^{k}+v_{id}^{k+1} \\
		v_{id}^{k+1}=wv_{id}^{k} + c_1r_1(p_{id, pbest}^k-x_{id}^k) + \\ c_2r_2(p_{d, gbest}^k-x_{id}^k)
	\end{aligned}
\end{equation}
$i$ denotes the $i^{th}$ individual, $d$ denotes in the $d^{th}$ dimension, $k$ denotes in the $k^{th}$ generation. In the velocity update equation, $w$ is a inertia weight, $c_1$ and $c_2$ are learning factor, $r_1$ and $r_2$ are random number.
 
\subsection{Estimation of a convergence point}
\begin{figure}[htb]
	\centering
	\includegraphics[width=9cm]{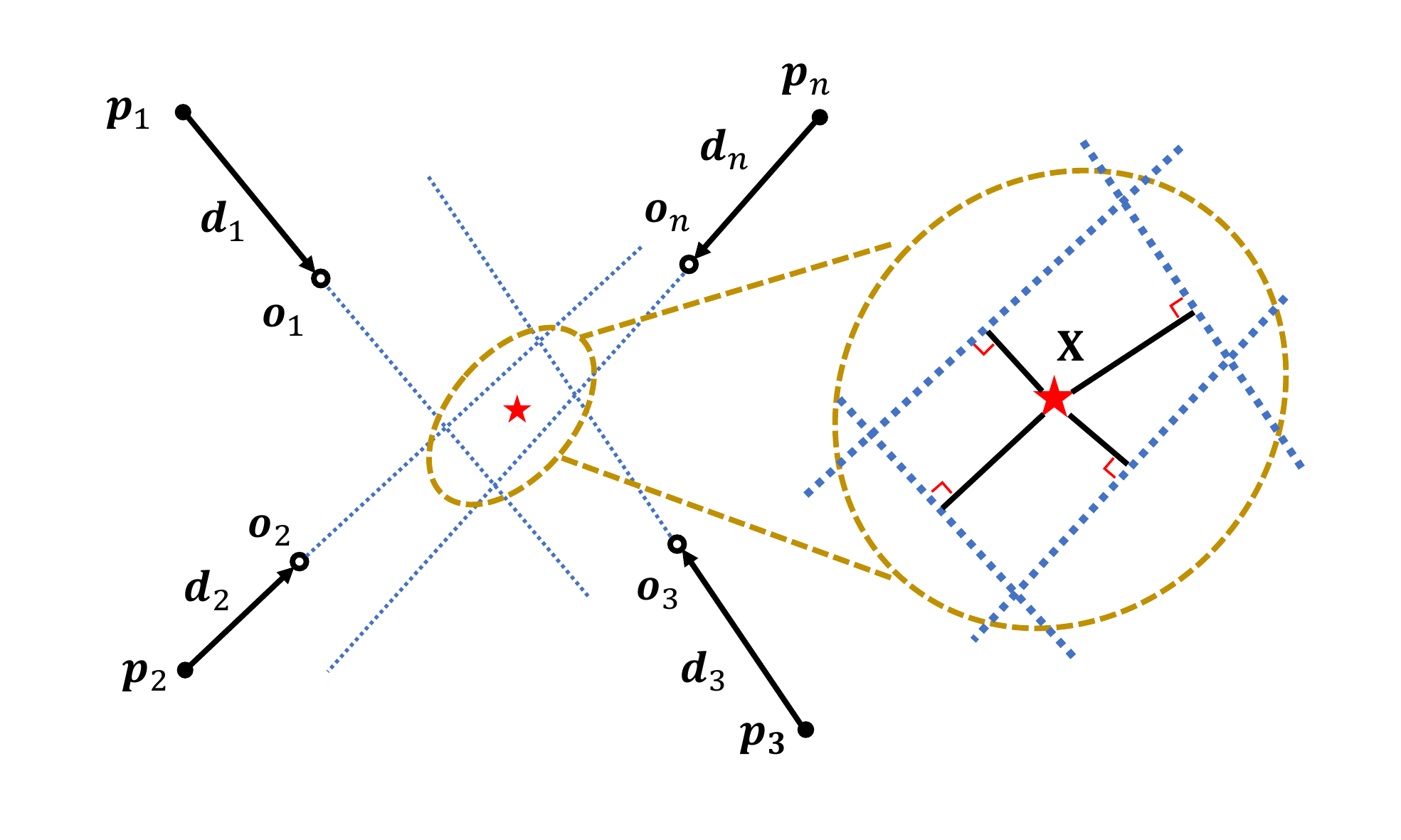}
	\caption{Moving vector $\boldsymbol{d}_i (=\boldsymbol{o}_i-\boldsymbol{p}_i)$ is calculated from a parent (worse) individual $\boldsymbol{p}_i$ and its offspring (better) $\boldsymbol{o}_i$. The $\star$ is the estimated convergence point.}
	\label{fig:1}
\end{figure}

The concept of estimation of a convergence point was first proposed by Murata. The paper\cite{Noboru:15} hypothesize that: in a population-based optimization algorithm, all individuals are moving towards the global optimal solution. According to this hypothesis, the nearest point to the extensions of all their movement directions should locate near the global optimal point. Fig.\ref{fig:1} shows how the estimation of a convergence point works. But in practice, the estimated point is not exactly on the global optimum due to incorrect directions or inaccurate directions of population movements. However, it is highly expected that the estimation point is close to the global optimum at least for unimodal functions.

Let us derive how to estimate the convergence point mathematically. First, we define parent (worse) individual $\boldsymbol{p}_i$, offspring (better) $\boldsymbol{o}_i$, and moving vector $\boldsymbol{d}_i$ as describe in Fig \ref{fig:1}. The unit direction vector of $\boldsymbol{d}_i$ is given as $\boldsymbol{d}_{0i}=\frac{\boldsymbol{d}_i}{||\boldsymbol{d}_i||}$,i.e., $\boldsymbol{d}_{0i}^{\rm T}\boldsymbol{d}_{0i}=1$. $\mathbf{X}$ denotes the estimated convergence point, and $\boldsymbol{p}_i+t_i\boldsymbol{d}_i$ represents the expansion from parent individual $\boldsymbol{p}_i$ with the direction $\boldsymbol{d}_i$. $L(\mathbf{X},{t_i})$ in Eq (\ref{eq:1}) becomes the minimum.
\begin{equation}
	\begin{aligned}
		\label{eq:1}
		\min (L(\mathbf{X},{t_i})) = \min (\sum_{i=1}^{n}||\boldsymbol{p}_i+t_i\boldsymbol{d}_i-\mathbf{X}||^2)
	\end{aligned}
\end{equation}

As the minimum line segment from the convergence point $\mathbf{X}$ to the expansion line segments is the orthogonal projection from $\mathbf{X}$, we can apply the Eq (\ref{eq:2}) into Eq (\ref{eq:1}) to remote $t_i$.
\begin{equation}
	\begin{aligned}
		\label{eq:2}
		\boldsymbol{d}_i^T(\boldsymbol{p}_i+t_i\boldsymbol{d}_i-\mathbf{X})=0 \ ({\rm orthogonal \ condition})
	\end{aligned}
\end{equation}

Finally, the convergence point $\mathbf{X}$ can be calculated by Eq (\ref{eq:3}). See detail expansion of equations in paper \cite{Noboru:15}.
\begin{equation}
	\begin{aligned}
		\label{eq:3}
		\widehat{\mathbf{X}}=\left\lbrace \sum_{i=1}^{n}(\boldsymbol{I}_d-\boldsymbol{d}_{0i}\boldsymbol{d}_{0i}^{\rm T}) \right\rbrace ^{-1}\left\lbrace \sum_{i=1}^{n}(\boldsymbol{I}_d-\boldsymbol{d}_{0i}\boldsymbol{d}_{0i}^{\rm T})\boldsymbol{p}_i\right\rbrace
	\end{aligned}
\end{equation}

\section{Proposal} \label{sec:3}
\begin{figure}[htb]
	\centering
	\includegraphics[width=9cm]{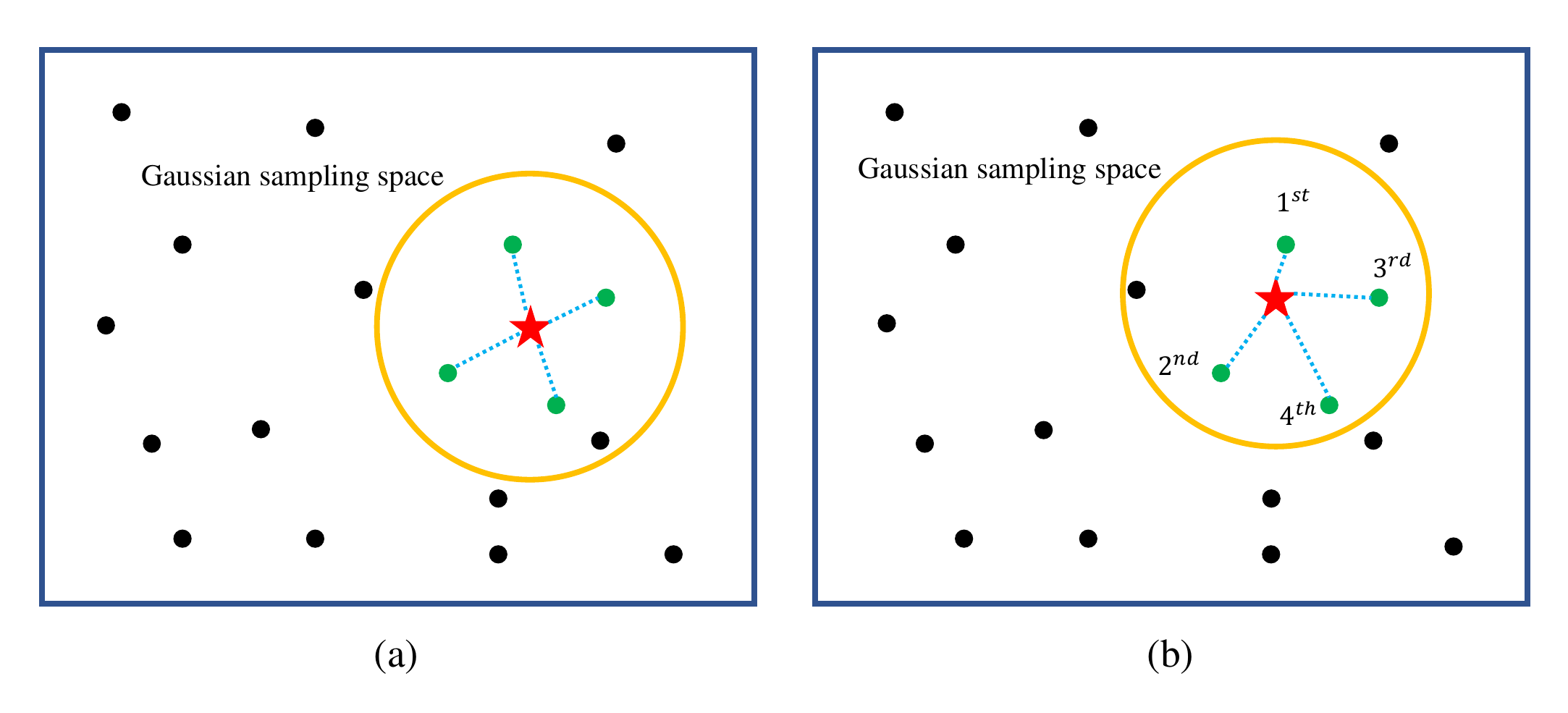}
	\caption{(a).elite sub-population averaging strategy for approximating the convergence point. (b).elite sub-population weighted averaging strategy for approximating the convergence point. The orange circle is the Gaussian sampling space with the mean of estimated convergence point.}
	\label{fig:2}
\end{figure}

In this Section, we will introduce our proposal in detail. First, we propose two approximate methods for estimating the convergence points with Gaussian sampling, which are shown in Fig \ref{fig:2}

First, we proposed a hypothesis that there is a higher probability for a optimum exists around an elite sub-population (individuals with higher fitness). Based on this hypothesis, we select the elite sub-population from the individual population of each generation from DE, and apply the averaging strategy and the weighted averaging strategy to approximately estimate the convergence point. Algorithm \ref{alg:6} and Algorithm \ref{alg:7} show the procedure of averaging strategy and weighted averaging strategy.
\begin{algorithm}
	\label{alg:6}
	\DontPrintSemicolon
	\SetAlgoLined
	\KwIn {${\rm Population}:P;{\rm Elite \ rate}:r;{\rm Dimension}:D$}
	\KwOut {${\rm Estimated \ convergence \ point}: C$}
	\SetKwFunction{FAS}{\textbf{AS}}
	\SetKwProg{Fn}{Function}{:}{}
	\Fn{\FAS{$P, r$}}{
		$EP \gets \textbf{bestIndividuals}(P, r)$\;
		$s \gets \textbf{size}(EP)$\;
		\For{$i=0 \ to \ D$}{
			$C_i \gets 0$\;
			\For{$j=0 \ to \ s$}{
				$C_i \gets C_i + EP_{i,j}$\;
			}
			$C_i \gets C_i/s$\;
		}
		$\textbf{return} \ C$
	}
	\caption{Averaging strategy}
\end{algorithm}

\begin{algorithm}
	\label{alg:7}
	\DontPrintSemicolon
	\SetAlgoLined
	\KwIn {${\rm Population}:P;{\rm Elite \ rate}:r;{\rm Dimension}:D$}
	\KwOut {${\rm Estimated \ convergence \ point}: C$}
	\SetKwFunction{FWAS}{\textbf{WAS}}
	\SetKwProg{Fn}{Function}{:}{}
	\Fn{\FWAS{$P, r$}}{
		$EP \gets \textbf{bestIndividuals}(P, r)$\;
		$EF \gets \textbf{Evaluate}(EP)$\;
		$s \gets \textbf{size}(EP)$\;
		$F \gets 0$\;
		\For{$i=0 \ to \ s$}{
			$F \gets F+EF_i$\;
		}
		$\blacktriangleright$ (Calculate the weights) \;
		\For{$i=0 \ to \ s$}{
			$w_i \gets EF_i/F$\;
		}
		\For{$i=0 \ to \ D$}{
			$C_i \gets 0$\;
			\For{$j=0 \ to \ s$}{
				$C_i \gets C_i + w_j * EP_{i,j}$\;
			}
		}
		$\textbf{return} \ C$
	}
	\caption{Weighted averaging strategy}
\end{algorithm}
In Algorithm \ref{alg:7}, we calculate the weights based on normalized fitness of individuals. After the convergence point is estimated, we apply a Gaussian sampling operator with the mean of the convergence point and a certain standard deviation $\sigma$ to find $k$ individuals. Then we select the best $k$ individuals among 1 estimated convergence point and $k$ individuals found by Gaussian sampling operator and worst $k$ individuals generated by DE as a part of offspring to participate the optimization.

\section{Numerical Experiment} \label{sec:4}
Here, we define shortened names of our proposal:

(a) P1: proposal with averaging strategy and Gaussian sampling operator.

(b) P2: proposal with weighted averaging strategy and Gaussian sampling operator.

We apply 28 benchmark functions from the CEC2013 Suite\cite{Liang:13} with independent 30 trial runs in this evaluation experiment. Table \ref{tbl:1} shows the details of these functions.
\begin{table}[tbh]
	\scriptsize
	\centering
	\caption{CEC2013 Suite: Uni=unimodal, Multi=multimodal, Comp=Composition}
	\label{tbl:1}
	\begin{tabular}{cccc}
		\toprule
		Func. & Types & Characteristics & Optimum  \\
		\midrule
		$f_1$ & \multirow{5}*{Uni} & Sphere function & -1400   \\
		$f_2$ & ~ & Rotated high conditioned elliptic function & -1300   \\
		$f_3$ & ~ & Rotated Bent Cigar function & -1200   \\
		$f_4$ & ~ & Rotated discus function & -1100   \\
		$f_5$ & ~ & Different powers function & -1000   \\
		\midrule
		$f_6$ & \multirow{15}*{Multi} & Rotated Rosenbrockfs function & -900   \\
		$f_7$ & ~ & Rotated Schaffers function & -800   \\
		$f_8$ & ~ & Rotated Ackleyfs function & -700   \\
		$f_9$ & ~ & Rotated Weierstrass function  & -600   \\
		$f_{10}$ & ~ & Rotated Griewankfs function & -500   \\
		$f_{11}$ & ~ & Rastriginfs function  & -400   \\
		$f_{12}$ & ~ & Rotated Rastriginfs function & -300   \\
		$f_{13}$ & ~ & Non-continuous rotated Rastriginfs function & -200   \\
		$f_{14}$ & ~ & Schwefelfs function & -100   \\
		$f_{15}$ & ~ & Rotated Schwefelfs function & 100   \\
		$f_{16}$ & ~ & Rotated Katsuura function & 200   \\
		$f_{17}$ & ~ & Lunacek bi-Rastrigin function & 300   \\
		$f_{18}$ & ~ & Rotated Lunacek bi-Rastrigin function & 400   \\
		$f_{19}$ & ~ & Expanded Griewankfs plus Rosenbrockfs function & 500   \\
		$f_{20}$ & ~ & Expanded Schafferfs $f_6$ function & 600   \\
		\midrule
		$f_{21}$ & \multirow{8}*{Comp} & Composition function 1 ($n$ = 5, rotated) & 700   \\
		$f_{22}$ & ~ & Composition function 2 ($n$ = 3, unrotated) & 800   \\
		$f_{23}$ & ~ & Composition function 3 ($n$ = 3, rotated) & 900   \\
		$f_{24}$ & ~ & Composition function 4 ($n$ = 3, rotated) & 1000   \\
		$f_{25}$ & ~ & Composition function 5 ($n$ = 3, rotated) & 1100   \\
		$f_{26}$ & ~ & Composition function 6 ($n$ = 5, rotated) & 1200   \\
		$f_{27}$ & ~ & Composition function 7 ($n$ = 5, rotated) & 1300   \\
		$f_{28}$ & ~ & Composition function 8 ($n$ = 5, rotated) & 1400   \\
		\bottomrule
	\end{tabular}
\end{table}
And we compare our proposal with RS, GA, DE, ES and PSO. The parameter of these algorithm is shown in Table \ref{tbl:2}
\begin{table}[tbh]
	\scriptsize
	\centering
	\caption{The parameter of comparison methods}
	\label{tbl:2}
	\begin{tabular}{ccc}
		\toprule
		Alg. & Parameter & Value   \\
		\midrule
		\multirow{4}*{Common parameters} & dimension & 2-D, 10-D, 30-D \\
		~ & population size & 50 * dimension\\
		~ & max evaluation times &  1000 * dimension\\
		~ & search space & [-100, 100] * dimension \\
		\midrule
		RS & sampling strategy & random sampling   \\
		\midrule
		\multirow{3}*{GA} & strategy & Elitist GA   \\
		~ & crossover rate & 0.5   \\
		~ & mutation rate & 0.1   \\
		\midrule
		\multirow{3}*{DE} & strategy & DE/current-to-best/1   \\
		~ & scale factor & 0.7   \\
		~ & crossover rate & 0.9   \\
		\midrule
		\multirow{2}*{ES} & strategy & (1+1)-ES   \\
		~ & mutation strength & 0.2   \\
		\midrule
		\multirow{2}*{PSO} & $w$ & 0.9   \\
		~ & $c_1,c_2$ & 2   \\
		\midrule
		\multirow{5}*{Proposal} & strategy & DE/current-to-best/1   \\
		~ & scale factor & 0.7   \\
		~ & crossover rate & 0.9   \\
		~ & elite proportion & 0.05 \\
		~ & $\sigma$ in Gaussian sampling & 5 \\
		\bottomrule
	\end{tabular}
\end{table}

At the end of optimization, we apply the Kruskal-Wallis test and the Holm multiple comparison test on the fitness values. Table \ref{tbl:3} shows their results of the statistical tests. In Holm multiple comparison test, if our proposal is significant better than the second-best algorithm or original DE, we apply $\gg$ (p$<$0.01) and $>$ (p$<$0.05) to denote the significance, $\approx$ represents there is no significance between two comparing methods.
\begin{sidewaystable}[tbh]
	\scriptsize
	\centering
	\caption{CEC2013 Suite: Uni=unimodal, Multi=multimodal, Comp=Composition}
	\label{tbl:3}
	\begin{tabular}{cccccccccccccccc}
		\toprule
		\multirow{2}{*}{Func.} & \multicolumn{5}{c}{2-D} &  \multicolumn{5}{c}{10-D} & \multicolumn{5}{c}{30-D}  \\
		\cmidrule(r){2-6} \cmidrule(r){7-11} \cmidrule(r){12-16}
		~ & P1 & P2 & P1 with DE & P2 with DE & P1 with P2 & P1 & P2 & P1 with DE & P2 with DE & P1 with P2 & P1 & P2 & P1 with DE & P2 with DE & P1 with P2\\
		\midrule 
		$f_1$ & P1 $\gg$ PSO & P2 $\gg$ PSO & P1 $\gg$ DE & P2 $\gg$ DE & P1 $>$ P2 & P1 $\approx$ PSO & P2 $\approx$ PSO & P1 $\gg$ DE & P2 $\gg$ DE & P1 $\approx$ P2 & PSO $\approx$ P1 & PSO $\approx$ P2 & P1 $\gg$ DE & P2 $\gg$ DE & P1 $\approx$ P2  \\
		$f_2$ & P1 $\gg$ DE & P2 $\gg$ DE & P1 $\gg$ DE & P2 $\gg$ DE & P2 $\gg$ P1 & PSO $\gg$ P1 & PSO $\gg$ P2 & P1 $\gg$ DE & P2 $\gg$ DE & P1 $\approx$ P2 & PSO $\approx$ P1 & PSO $\approx$ P2 & P1 $\gg$ DE & P2 $\gg$ DE & P1 $\approx$ P2  \\
		$f_3$ & P1 $\gg$ DE & P2 $\gg$ DE & P1 $\gg$ DE & P2 $\gg$ DE & P1 $\approx$ P2 & PSO $\gg$ P1 & PSO $\gg$ P2 & P1 $\gg$ DE & P2 $\gg$ DE & P1 $\approx$ P2 & PSO $\approx$ P1 & PSO $\approx$ P2 & P1 $\gg$ DE & P2 $\gg$ DE & P1 $\approx$ P2  \\
		$f_4$ & P1 $\gg$ DE & P2 $\gg$ DE & P1 $\gg$ DE & P2 $\gg$ DE & P1 $\approx$ P2 & PSO $\gg$ P1 & PSO $\gg$ P2 & P1 $>$ DE & P2 $>$ DE & P1 $\approx$ P2 & PSO $\approx$ P1 & PSO $\approx$ P2 & P1 $\gg$ DE & P2 $\gg$ DE & P1 $\approx$ P2  \\
		$f_5$ & P1 $\gg$ PSO & P2 $\gg$ PSO & P1 $\gg$ DE & P2 $\gg$ DE & P1 $\approx$ P2 & PSO $\gg$ P1 & PSO $\gg$ P2 & P1 $>$ DE & P2 $>$ DE & P1 $\approx$ P2 & PSO $\approx$ P1 & PSO $\approx$ P2 & P1 $\gg$ DE & P2 $\gg$ DE & P1 $\approx$ P2  \\
		$f_6$ & P1 $\gg$ PSO & P2 $\gg$ PSO & P1 $\gg$ DE & P2 $\gg$ DE & P1 $\approx$ P2 & PSO $\gg$ P1 & PSO $\gg$ P2 & P1 $>$ DE & P2 $>$ DE & P1 $\approx$ P2 & PSO $\approx$ P1 & PSO $\approx$ P2 & P1 $\gg$ DE & P2 $\gg$ DE & P1 $\approx$ P2  \\
		$f_7$ & P1 $\approx$ PSO & P2 $\approx$ PSO & P1 $\gg$ DE & P2 $\gg$ DE & P1 $\approx$ P2 & PSO $\gg$ P1 & PSO $\gg$ P2 & P1 $\gg$ DE & P2 $\gg$ DE & P1 $\approx$ P2 & PSO $\approx$ P1 & PSO $\approx$ P2 & P1 $\gg$ DE & P2 $\gg$ DE & P1 $\approx$ P2  \\
		$f_8$ & P1 $\approx$ PSO & P2 $\approx$ PSO & P1 $\gg$ DE & P2 $\gg$ DE & P1 $\approx$ P2 & P1 $\approx$ PSO & P2 $\approx$ PSO & P1 $\approx$ DE & P2 $\approx$ DE & P1 $\approx$ P2 & P1 $\approx$ DE & P2 $\approx$ DE & P1 $\approx$ DE & P2 $\approx$ DE & P1 $\approx$ P2  \\
		$f_9$ & P1 $\approx$ PSO & P2 $\approx$ PSO & P1 $\gg$ DE & P2 $\gg$ DE & P1 $\approx$ P2 &  PSO $\gg$ P1 &  PSO $\gg$ P2 & P1 $>$ DE & P2 $\gg$ DE & P1 $\approx$ P2 & PSO $\approx$ P1 & PSO $\approx$ P2 & P1 $\approx$ DE & P2 $\approx$ DE & P1 $\approx$ P2  \\
		$f_{10}$ & P1 $\gg$ DE & P2 $\gg$ DE & P1 $\gg$ DE & P2 $\gg$ DE & P1 $\approx$ P2 &  PSO $\gg$ P1 &  PSO $\gg$ P2 & P1 $\gg$ DE & P2 $\gg$ DE & P1 $\approx$ P2 & PSO $\approx$ P1 & PSO $\approx$ P2 & P1 $\gg$ DE & P2 $\gg$ DE & P1 $\approx$ P2  \\
		$f_{11}$ & P1 $\approx$ PSO & P2 $>$ PSO & P1 $\gg$ DE & P2 $\gg$ DE & P1 $\approx$ P2 & P1 $\approx$ GA & P2 $\approx$ GA & P1 $\gg$ DE & P2 $\gg$ DE & P1 $\approx$ P2 & PSO $\gg$ P1 & PSO $\gg$ P2 & P1 $\gg$ DE & P2 $\gg$ DE & P1 $\approx$ P2  \\
		$f_{12}$ & P1 $\approx$ PSO & P2 $\approx$ PSO & P1 $\gg$ DE & P2 $\gg$ DE & P1 $\approx$ P2 & P1 $\gg$ PSO & P2 $\gg$ PSO & P1 $\gg$ DE & P2 $\gg$ DE & P1 $\approx$ P2 & PSO $\gg$ P1 & PSO $\gg$ P2 & P1 $\gg$ DE & P2 $\gg$ DE & P1 $\approx$ P2  \\
		$f_{13}$ & P1 $\approx$ PSO & P2 $\approx$ PSO & P1 $\gg$ DE & P2 $\gg$ DE & P1 $\approx$ P2 & P1 $\gg$ PSO & P2 $\gg$ PSO & P1 $\gg$ DE & P2 $\gg$ DE & P1 $\approx$ P2 & PSO $\gg$ P1 & PSO $\gg$ P2 & P1 $\gg$ DE & P2 $\gg$ DE & P1 $\approx$ P2  \\
		$f_{14}$ & P1 $\approx$ GA & P2 $\approx$ GA & P1 $\approx$ DE & P2 $\gg$ DE & P1 $\approx$ P2 & P1 $\gg$ PSO & P2 $\gg$ PSO & P1 $\approx$ DE & P2 $\approx$ DE & P1 $\approx$ P2 & GA $\gg$ P1 & GA $\gg$ P2 & P1 $\approx$ DE & P2 $>$ DE & P1 $\approx$ P2  \\
		$f_{15}$ & P1 $\approx$ PSO & P2 $\approx$ PSO & P1 $\gg$ DE & P2 $>$ DE & P1 $\approx$ P2 & P1 $\gg$ PSO & P2 $\gg$ PSO & P1 $\gg$ DE & P2 $\gg$ DE & P1 $\approx$ P2 & P1 $\gg$ PSO & P2 $\gg$ PSO & P1 $\gg$ DE & P2 $\gg$ DE & P1 $\approx$ P2  \\
		$f_{16}$ & P1 $\approx$ DE & P2 $\approx$ DE & P1 $\approx$ DE & P2 $\approx$ DE & P1 $\approx$ P2 & P1 $\approx$ GA & P2 $\approx$ GA & P1 $\approx$ DE & P2 $\approx$ DE & P1 $\approx$ P2 & P1 $\approx$ ES & P2 $\approx$ ES & P1 $\approx$ DE & P2 $\approx$ DE & P1 $\approx$ P2  \\
		$f_{17}$ & P1 $\approx$ DE & P2 $\approx$ DE & P1 $\approx$ DE & P2 $\approx$ DE & P1 $\approx$ P2 & P1 $\gg$ PSO & P2 $\gg$ PSO & P1 $\approx$ DE & P2 $\approx$ DE & P1 $\approx$ P2 & P1 $\gg$ PSO & P2 $\gg$ PSO & P1 $\approx$ DE & P2 $\approx$ DE & P1 $\approx$ P2  \\
		$f_{18}$ & P1 $\approx$ PSO & P2 $\approx$ PSO & P1 $\approx$ DE & P2 $\approx$ DE & P1 $\approx$ P2 & P1 $\gg$ PSO & P2 $\gg$ PSO & P1 $\gg$ DE & P2 $\gg$ DE & P1 $\approx$ P2 & PSO $\gg$ P1 & PSO $>$ P2 & P1 $\gg$ DE & P2 $\gg$ DE & P1 $\approx$ P2  \\
		$f_{19}$ & P1 $\approx$ PSO & P2 $\approx$ PSO & P1 $\approx$ DE & P2 $\approx$ DE & P1 $\approx$ P2 & P1 $\gg$ PSO & P2 $>$ PSO & P1 $\gg$ DE & P2 $\gg$ DE & P1 $\approx$ P2 & PSO $\gg$ P1 & PSO $\gg$ P2 & P1 $\gg$ DE & P2 $\gg$ DE & P1 $\approx$ P2  \\
		$f_{20}$ & P1 $\approx$ PSO & P2 $\approx$ PSO & P1 $\gg$ DE & P2 $\gg$ DE & P1 $\approx$ P2 & PSO $\gg$ P1 & PSO $\gg$ P2 & P1 $\gg$ DE & P2 $\gg$ DE & P1 $\approx$ P2 & PSO $\gg$ P1 & PSO $\gg$ P2 & P1 $>$ DE & P2 $>$ DE & P1 $\approx$ P2  \\
		$f_{21}$ & P1 $\gg$ PSO & P2 $\gg$ PSO & P1 $\gg$ DE & P2 $\gg$ DE & P1 $\approx$ P2 & PSO $\gg$ P1 & PSO $>$ P2 & P1 $\gg$ DE & P2 $\gg$ DE & P1 $\approx$ P2 & PSO $\gg$ P1 & PSO $\gg$ P2 & P1 $\gg$ DE & P2 $\gg$ DE & P1 $\approx$ P2  \\
		$f_{22}$ & P1 $\approx$ GA & P2 $\approx$ GA & P1 $\gg$ DE & P2 $\gg$ DE & P1 $>$ P2 & PSO $\gg$ P1 & PSO $\gg$ P2 & P1 $\gg$ DE & P2 $\gg$ DE & P1 $\approx$ P2 & GA $\gg$ P1 & GA $\gg$ P2 & P1 $\gg$ DE & P2 $\gg$ DE & P1 $\approx$ P2  \\
		$f_{23}$ & P1 $\approx$ PSO & P2 $\approx$ PSO & P1 $\gg$ DE & P2 $\gg$ DE & P1 $\approx$ P2 & P1 $\approx$ PSO & P2 $\approx$ PSO & P1 $\gg$ DE & P2 $\gg$ DE & P1 $\approx$ P2 & P1 $\approx$ PSO & P2 $\approx$ PSO & P1 $\approx$ DE & P2 $\approx$ DE & P1 $\approx$ P2  \\
		$f_{24}$ & P1 $\approx$ PSO & P2 $\approx$ PSO & P1 $\gg$ DE & P2 $\gg$ DE & P1 $\approx$ P2 & P1 $\approx$ ES & P2 $\approx$ ES & P1 $\gg$ DE & P2 $\gg$ DE & P1 $\approx$ P2 & PSO $\gg$ P1 & PSO $\gg$ P2 & P1 $\gg$  DE & P2 $\gg$  DE & P1 $\approx$ P2  \\
		$f_{25}$ & P1 $\approx$ PSO & P2 $\approx$ PSO & P1 $\gg$ DE & P2 $\gg$ DE & P1 $\approx$ P2 & P1 $\approx$ PSO & P2 $\approx$ PSO & P1 $>$ DE & P2 $\gg$ DE & P1 $\approx$ P2 & PSO $\gg$ P1 & PSO $\gg$ P2 & P1 $\approx$ DE & P2 $\approx$ DE & P1 $\approx$ P2  \\
		$f_{26}$ & P1 $\approx$ PSO & P2 $\approx$ PSO & P1 $\gg$ DE & P2 $\gg$ DE & P1 $\approx$ P2 & P1 $\approx$ GA & P2 $>$ GA & P1 $\gg$ DE & P2 $\gg$ DE & P1 $\approx$ P2 & PSO $\gg$ P1 & PSO $\gg$ P2 & P1 $\approx$ DE & P2 $\gg$ DE & P1 $\approx$ P2  \\
		$f_{27}$ & P1 $\approx$ PSO & P2 $\approx$ PSO & P1 $\approx$ DE & P2 $\approx$ DE & P1 $\approx$ P2 & PSO $\gg$ P1 & PSO $\gg$ P2 & P1 $\gg$ DE & P2 $\gg$ DE & P1 $\approx$ P2 & PSO $\gg$ P1 & PSO $\gg$ P2 & P1 $\approx$ DE & P2 $\gg$ DE & P1 $\approx$ P2  \\
		$f_{28}$ & P1 $\approx$ PSO & P2 $\approx$ PSO & P1 $\approx$ DE & P2 $\approx$ DE & P1 $\approx$ P2 & P1 $\approx$ PSO & P2 $\approx$ PSO & P1 $>$ DE & P2 $\gg$ DE & P1 $\approx$ P2 & PSO $\gg$ P1 & PSO $\gg$ P2 & P1 $\approx$ DE & P2 $\gg$ DE & P1 $\approx$ P2  \\
		\bottomrule
	\end{tabular}
\end{sidewaystable}

From the experiment results, we can see that our proposal can significantly accelerate the DE in most test functions on the CEC2013 Suite, and P1 has no siginificant difference with P2 in most functions. In the 2-D space, our proposal is competitive with the comparing 5 methods, while in the 10-D and 50-D space, although PSO outperforms our proposal combined with DE on most test functions, this is mainly due to the limited performance of DE on these functions, our proposal can still accelerate the DE in optimization. Different from the method proposed by Yu et al\cite{Yu:19}, it is unnecessary for our proposal to calculate the moving vector, so it has stronger scalability for our proposal combined with other evolutionary algorithms.

\section{Conclusion} \label{sec:5}
In this paper, we propose a simple strategy to estimate the convergence point approximately. Based on the averaging strategy, ordinary averaging strategy and weighted averaging strategy are derived. We also design a Gaussian sampling operator based on the estimated convergence point and combine it with traditional DE. Numerical experiments show our proposal can improve the performance of traditional DE. Although our proposal combined with DE is worse than PSO in some functions with 10-D and 50-D, the main reason is the performance of DE is limited in these functions, and our proposal has strong scalability that can be extended to all population-based heuristic algorithms in theory. 

In future research, a well-performed local search operator replacing the Gaussian sampling is a promising topic, such as the introduction of CMA-ES or the adaptive evolution strategy. And in this paper, we only apply our proposal in low dimensional test functions. In future research, we will combine our proposal with cooperative co-evolution to solve large-scale optimization problems. Finally, the estimation of a convergence point combined with the Gaussian sampling operator has a broad prospect to solve optimization problems and can be easily extended with other population-based evolutionary algorithms.

\section{Acknowledgement}
This work was supported by JSPS KAKENHI Grant Number JP20K11967.

\bibliographystyle{IEEEtran}
\bibliography{Paper}

\end{document}